\documentclass[journal]{IEEEtran}

\usepackage{amsmath,graphicx,cite,url,pgfplots,pgfplotstable,booktabs,hhline,tabularx,hyperref}

\pgfplotsset{
  /pgfplots/my xbar legend/.style={
    /pgfplots/legend image code/.code={%
      \draw[##1,/tikz/.cd,bar width=.5em]
      plot coordinates { (\pgfplotbarwidth,0.1em)};}
}}

\makeatletter
\def\ps@IEEEtitlepagestyle{
  \def\@oddfoot{\mycopyrightnotice}
  \def\@evenfoot{}
}
\def\mycopyrightnotice{
  {\footnotesize
  \begin{minipage}{\textwidth}
  \centering
  Copyright~\copyright~2017 IEEE. Personal use of this material is permitted. However, permission to use this  \\ 
  material for any other purposes must be obtained from the IEEE by sending a request to pubs-permissions@ieee.org.
  \end{minipage}
  }
}

\begin{document}

\title{Multi-Level and Multi-Scale Feature Aggregation Using Pre-trained  Convolutional Neural Networks for Music Auto-tagging}

%

\author{Jongpil~Lee and Juhan~Nam,~\IEEEmembership{Member,~IEEE}

\thanks{J. Lee and J. Nam are with the Graduate School of Culture Technology, Korea Advanced Institute of Science and Technology (KAIST), South Korea.}

}
%
%

\markboth{IEEE SIGNAL PROCESSING LETTERS}%
{Lee and Nam: Multi-Level and Multi-Scale Feature Aggregation Using Pre-trained Convolutional Neural Networks for Music Auto-tagging}
%


\maketitle

\begin{abstract}

Music auto-tagging is often handled in a similar manner to image classification by regarding the 2D audio spectrogram as image data. However, music auto-tagging is distinguished from image classification in that the tags are highly diverse and have different levels of abstractions. Considering this issue, we propose a convolutional neural networks (CNN)-based architecture that embraces multi-level and multi-scaled features. The architecture is trained in three steps. First, we conduct supervised feature learning to capture local audio features using a set of CNNs with different input sizes. Second, we extract audio features from each layer of the pre-trained convolutional networks separately and aggregate them altogether given a long audio clip. Finally, we put them into fully-connected networks and make final predictions of the tags. Our experiments show that using the combination of multi-level and multi-scale features is highly effective in music auto-tagging and the proposed method outperforms previous state-of-the-arts on the MagnaTagATune dataset and the Million Song Dataset. We further show that the proposed architecture is useful in transfer learning.

\end{abstract}


\section{Introduction}
\label{sec:intro}
\IEEEPARstart{M}{u} sic auto-tagging is a task that predicts descriptive words of music from the audio signals. Recently, as convolutional neural networks (CNN) has become the de-facto standard in image classification, the deep learning approach has been actively explored in music auto-tagging as well by using the spectrogram and its variants as input data and so recasting it as a multi-label classification task on the 2-D time-frequency images \cite{P.Hamel:11,dieleman2014end,pons2016experimenting,choi2016automatic}. 

However, music auto-tagging is distinguished from image classification in that the tags consist of words that are highly diverse and have different levels of abstractions. For example, some words, particularly instrument-related ones, such as \textit{female vocalist}, \textit{guitar} and \textit{saxophone} are objective descriptions of specific sound sources. They tend to be local and repetitive within an audio clip and are basically predicted from the physical properties of the sound sources. On the other hand,  other words such as \textit{rock}, \textit{happy} and \textit{80s} are discriminative descriptions of the overall content in terms of genre, mood and years. They are more global and comprehensive, requiring longer audio segments to predict them.


While the tags are positioned in different levels or time-scales in a hierarchy, the majority of previous work predicted them from the same level or scale of features as typically done in image classification. There are a few that handled this issue explicitly by comparing or combining multi-layer or multi-scale audio features. In terms of feature level, Lee et. al. used convolutional restricted Boltzmann machine to learn hierarchical features in an unsupervised manner \cite{Lee+etal09:convDBNAudio}. They compared each layer of features and their combination for music genre and artist classification. However, the evaluation was not sufficiently comprehensive as they used small datasets. Hamel and Eck applied deep neural networks pre-trained with deep belief networks to learn hierarchical features and compared each layer of features \cite{hamel2010learning}. However, the learned features were obtained from single frames of spectrogram, which is too local to capture musically meaningful and rich patterns. In terms of time scale, Hamel et. al. investigated combining different resolutions of spectrograms \cite{P.Hamel:12}, and Dieleman and Schrauwen improved the approach further using Gaussian and Laplacian pyramids \cite{dieleman2013multiscale}. However, they conducted the multi-scaling and feature concatenation only on the input layer, focusing on the spectrograms. 


The general consensus from the previous work is that individual tags have different performance sensitivity to different time scales and levels of features, and so combining all of them provides the best results. With this lesson, we propose a CNN-based architecture that handles multi-level and multi-scale of audio features more comprehensively. The architecture is trained in three steps: local feature learning, feature aggregation and global classification. The local feature learning is carried out using a set of CNNs. They are trained in a supervised manner with the tag labels, taking different sizes of input frames and accordingly more hidden layers. The feature aggregation step extracts local features from all layers and time-scales using the pre-trained CNNs and summarizes them into a single feature vector. The last stage performs final predictions of tags from the aggregated feature vector using a fully-connected neural network. By nature of the separated steps, this architecture is capable of transfer learning, that is, by conducting local feaure learning with one large dataset and then feature aggregation and global classification with another dataset.   

We evaluate the proposed architecture with two popularly used datasets primarily for music auto-tagging and also in transfer learning settings where music genre classification is performed using pre-trained CNNs. Our experiments show how different combinations of multi-layer and multi-time-scale features improve the accuracy and also the architecture outperforms previous state-of-the-arts. 

\section{Proposed Method}
\label{sec:model}

\begin{figure}[t]
\vspace{-4mm}
\centering
\centerline{\includegraphics[width=9.2cm]{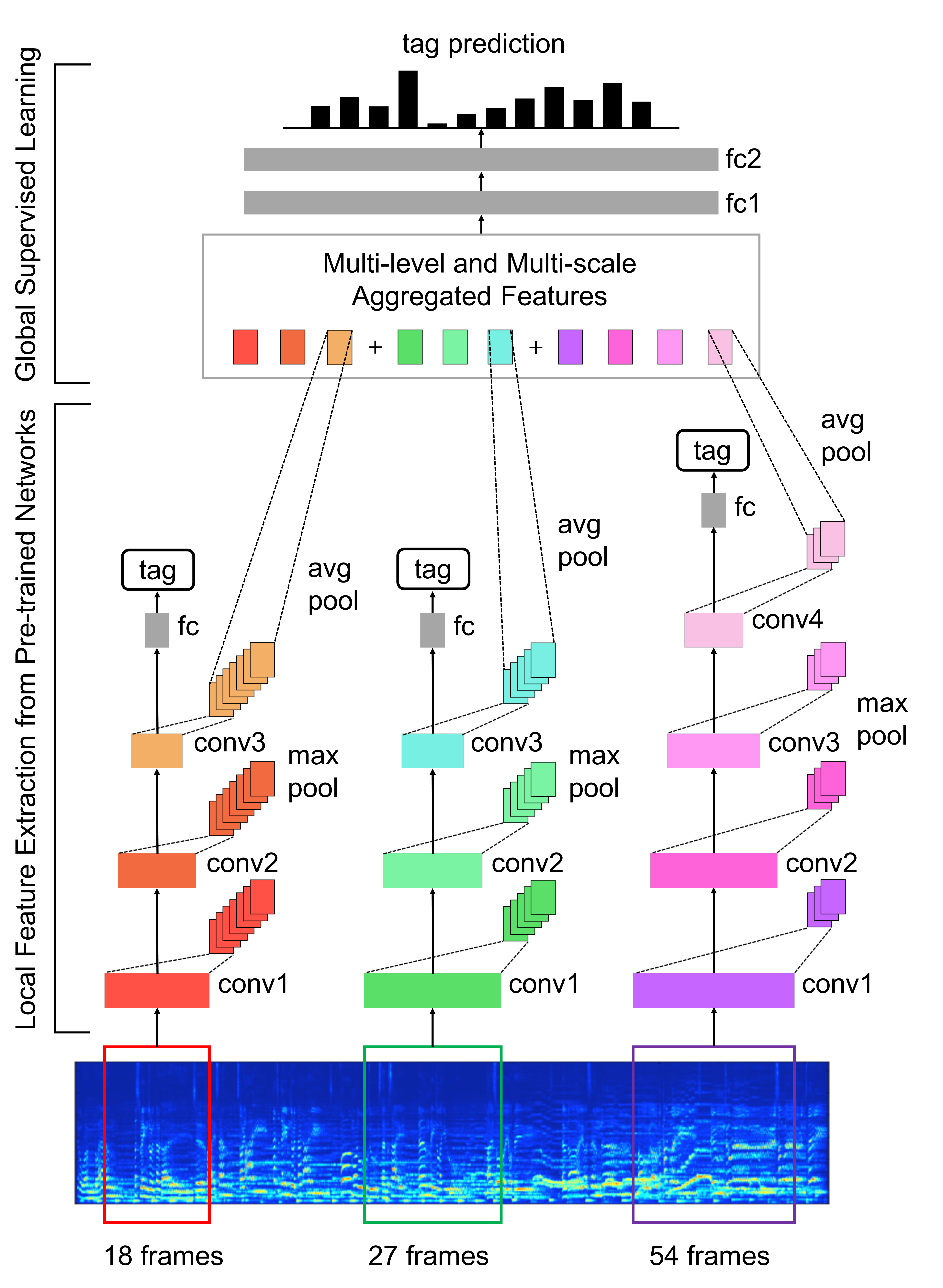}} 
\caption{The proposed architecture for music auto-tagging. "fc" stands for "fully-connected layer". The CNNs capture local audio features by supervised learning and are used as multi-level and multi-scale feature extractors. Given a long audio clip, the extracted features are aggregated. Finally, the global classifier makes final predictions of tags on them.}
\label{fig:fig1}
\end{figure}

The overall architecture that we propose is illustrated in Figure \ref{fig:fig1}. The followings describe the three steps to train it. 

\subsection{Local feature learning}
\label{Local feature learning}
In the first step, we perform supervised feature learning with a set of CNNs. We chose the segment sizes such that the hidden layers capture multi-level audio features within one to several musical beats for different beats per minute (BPM). For example, 18, 27 and 54 frames correspond to 420, 630 and 1260 msec. They take care of at least one beat long in songs with 48 to 143 BPM, which is the tempo range that covers the majority of popular music. The CNNs are configured to conduct 1-D convolution in all layers, assuming that low-frequency and high-frequency content do not share the weights (as opposed to images) and so the whole frequency range is under the receptive field of the filters. We determined the filter width "3" in the convolutional layers by referring to the VGG net \cite{simonyan2014very}. We first built 27 frames model, which is composed of 3 convolutional layers ((3,128)-(3)-(3,128)-(3)-(3,256)-(3), (filter length, number of filters)-(pool length)), 1 fully-connected layer (256), and final prediction layer (50). Using this as a reference configuration, For the 18 frames model, the last convolutional layer was replaced by (2,256)-(2) and for models with more than 54 frames as input, we simply added (2,256)-(2) layers according to the input size. Zero-padding is applied to each convolution layer to preserve the size. The convolution stride is fixed to 1 and max-pooling stride is set to the same as the pooling length. We perform the back-propagation with tag labels from the dataset. Each of the models can be actually used to predict the tags for a long audio clip by averaging the local predictions. We call them "local models"




\subsection{Multi-level and multi-scale feature aggregation}
The pre-trained CNNs can be viewed as feature extractors. Since a single CNN model can extract different levels of features given the input size and we train them with different input sizes, we can extract multi-level and multi-scale features from them. In order to handle the long audio clips (typically, about 30 secs), we aggregate them into a single large feature vector and use them as a song-level representation. For example, the output shape of each convolution layer of one segment in a 27 frames model is (27,128)-(9,128)-(3,256). After extracting features for all segments in 30 seconds audio, the song-level feature dimension become (46,27,128)-(46,9,128)-(46,3,256). In order to extract the most representative feature, we apply max-pooling over each segment separately. We then summarize them into single feature vectors by average pooling over the long audio clip separately for each layer. This scheme, that is, max-pooling followed by average pooling, was used as an effective means to summarize local features \cite{nam2012learning, name2015deepbof}. As a result, the dimensionality of the concatenated feature vector will be determined by the sum of the numbers of filters from all layers. For instance, the 27 frames model will have 128 + 128 + 256 dimensional feature vector. This is repeated for all different input sizes and they are finally concatenated into a large feature vector.



\subsection{Global classification}
In this step, we make final predictions of tags from the aggregated multi-level and multi-scale features. We train another classifier, which a neural network with two fully connected hidden layers with 512 or 1024 units depending on the datasets. Since the feature aggregation and global classification steps are separated from the local feature learning, we can conduct transfer learning, which has been explored effectively as well for music audio data\cite{Hamel2013, van2014transfer}, using pre-trained CNNs with a large dataset. In our experiment, we evaluate the transfer learning setting for several different datasets. 





\section{Experiments}

\subsection{Datasets}
To evaluate the proposed architecture, we primarily used the MagnaTagATune (MTAT) dataset \cite{law2009evaluation} and Million Song Dataset (MSD) annotated with the Last.fm tags \cite{bertin2011million}. We filtered out the tags and used most frequently labeled 50 tags for both MTAT and MSD, following the previous work\cite{dieleman2014end,choi2016automatic}\footnote{ \url{https://github.com/keunwoochoi/MSD_split_for_tagging}}. Also, all songs in the two datasets were trimmed to 29.1 second long. We used AUC (Area Under Receiver Operating Characteristic) as a primary evaluation metric for music auto-tagging. In addition, we conducted genre classification tasks, GTZAN\cite{tzanetakis2002musical} (10 genres, fault-filtered split that is designed to avoid the repetition of artist across training/validation/test list\cite{kereliuk2015deep}) and Tagtraum genre annotations on MSD (15 genres, stratified split with 80\% training data of CD2C version)\cite{schreiberimproving}, in a transfer learning setting where the pre-trained CNNs with MSD are used as feature extractors. 

\begin{figure}[t]
\vspace{-9mm}
\centering
\centerline{\includegraphics[width=8.6cm]{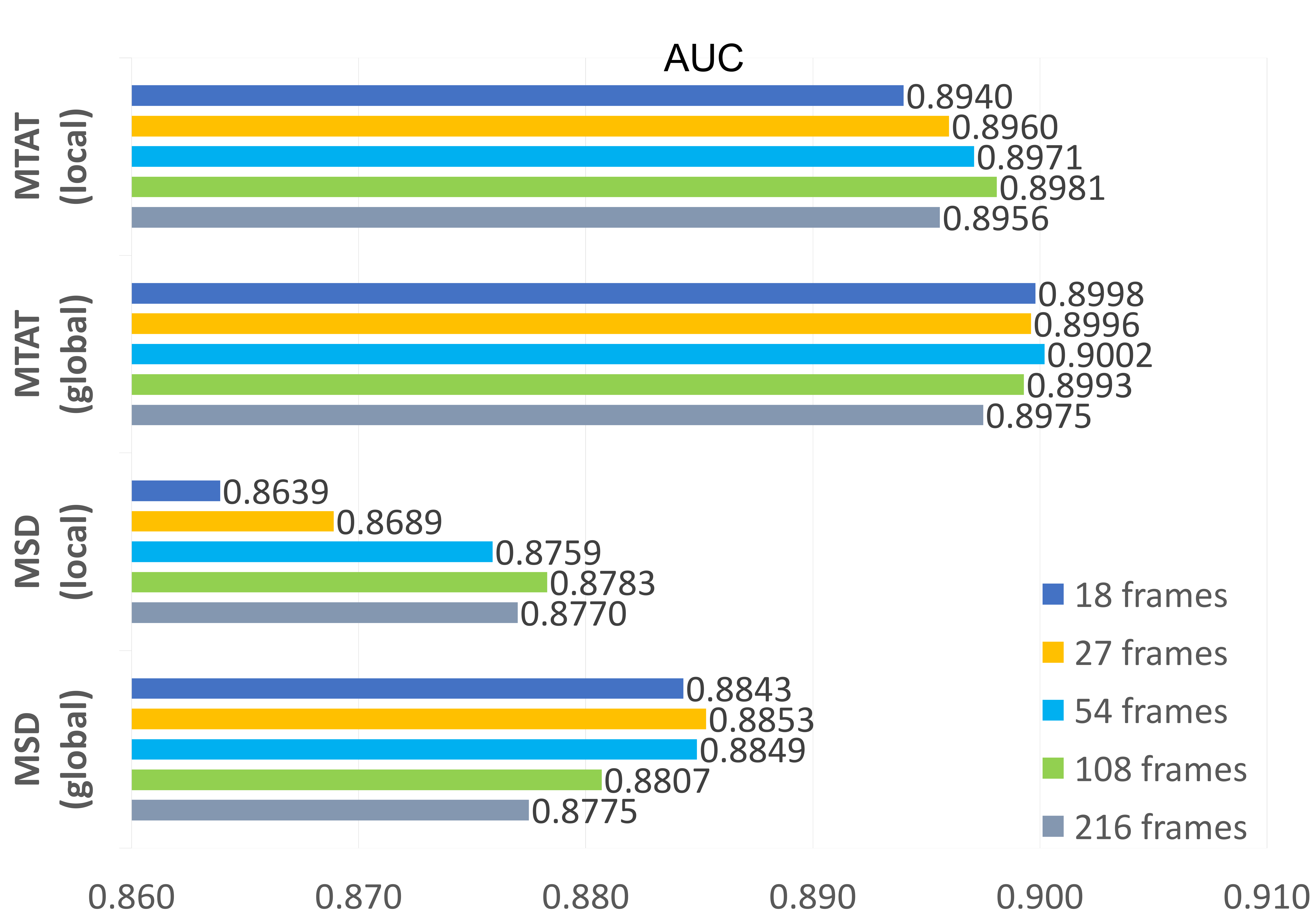}}
\caption{Comparison of local and global models for different input sizes.}
\label{fig:fig3}
\end{figure}

\begin{figure}[t]
\vspace{-4.6mm}
\centering
\centerline{\includegraphics[width=8.6cm]{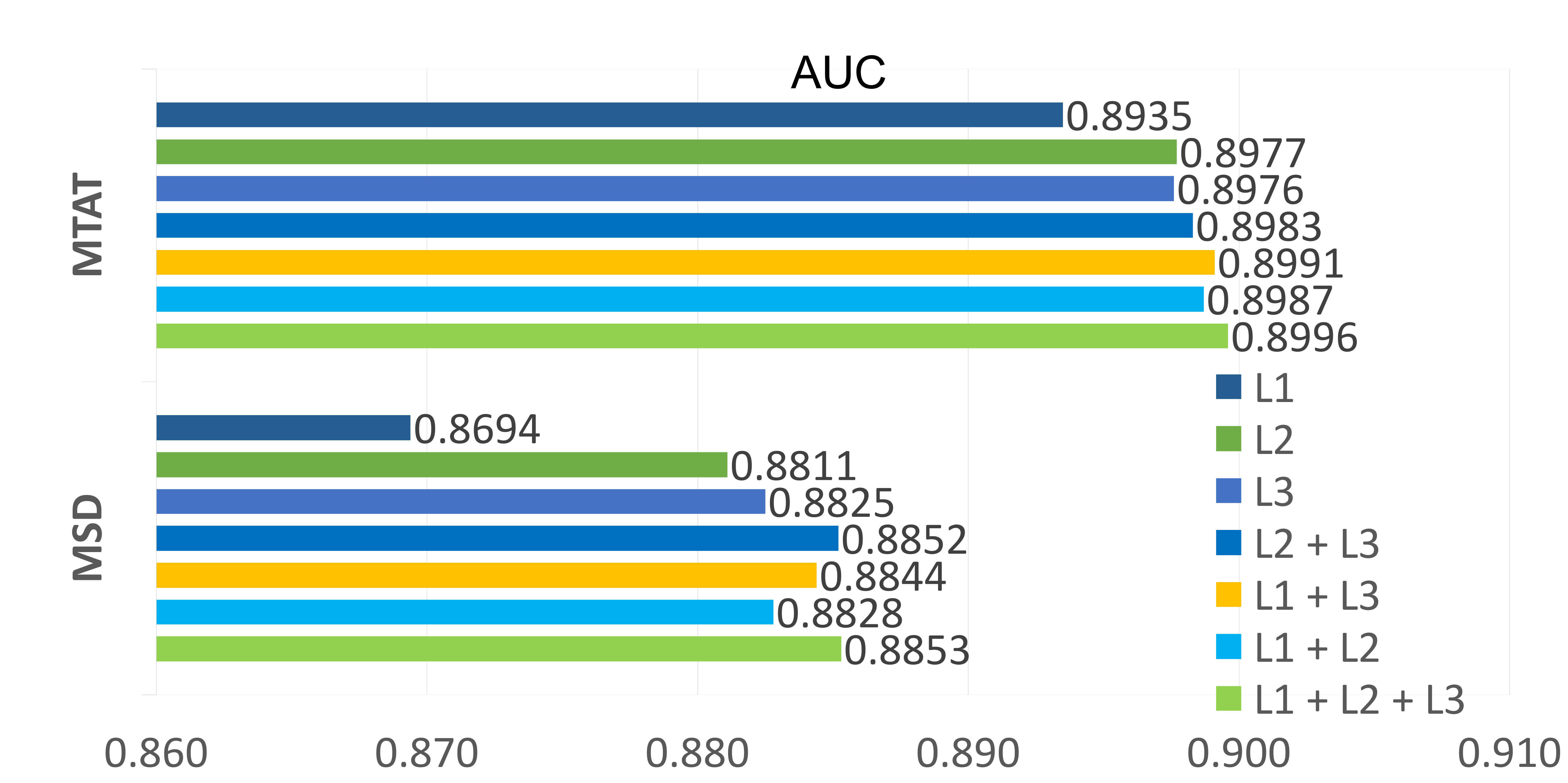}}
\caption{Comparison of various combinations of feature levels in the 27 frames model. L1, L2 and L3 denote the hidden layer activations in each layer in the CNN model.}
\label{fig:fig4}
\end{figure}

\begin{figure}[t]
\vspace{-8.5mm}
\centering
\centerline{\includegraphics[width=8.6cm]{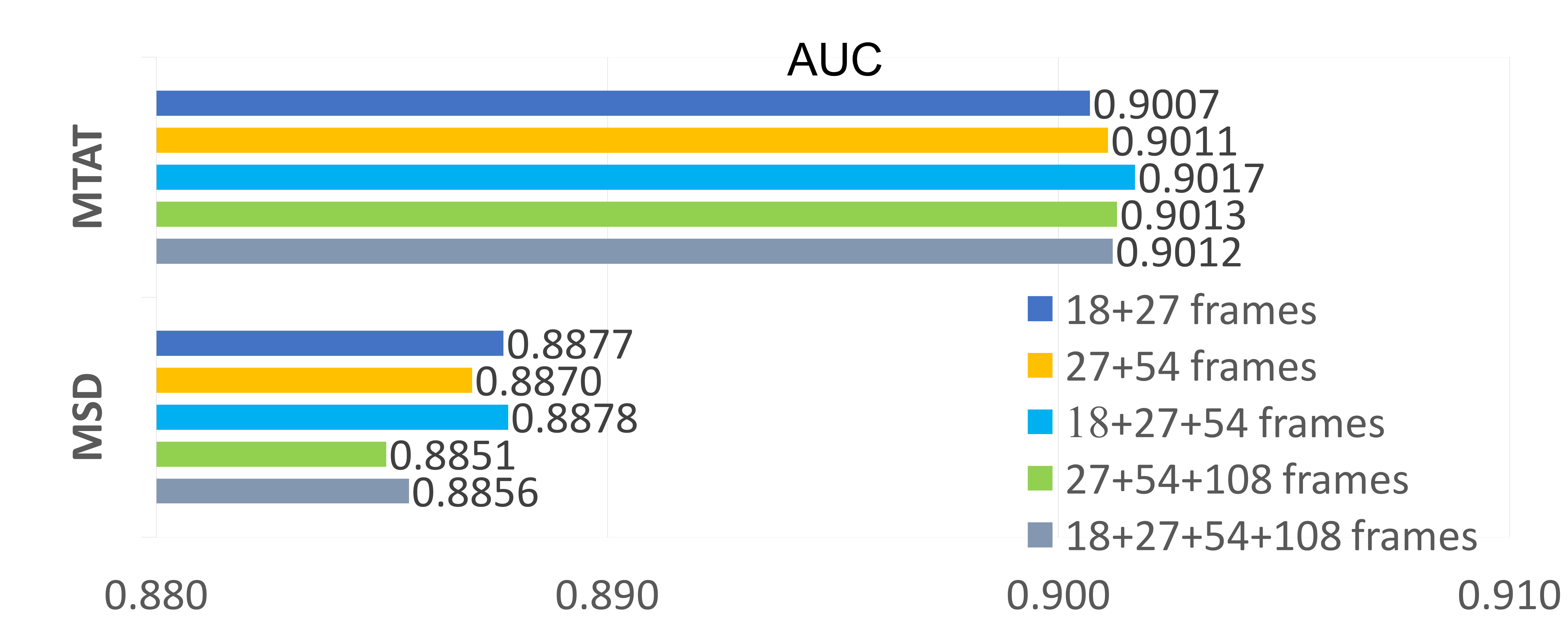}}
\caption{Comparison of various combinations of multi-scale features.}
\label{fig:fig5}
\end{figure}

\subsection{Training details}
Mel-frequency spectrogram with 128 bins are used as the input representation. The parameters are set to 22.05 kHz sampling rate (by resampling when necessary), 512 samples of hop size, 1024 samples of Hanning window, and magnitude compression with a nonlinear curve, $\log(1+C|A|)$ where $A$ is the magnitude and C is set to 10. As a result, each clip has 1250 frames and is divided into 69, 46, 23, 11 and 5 segments for the corresponding 18, 27, 54, 108 and 216 frames models, respectively. The detailed parameters to train the networks are as follows: sigmoid activation for output layer with binary cross entropy loss, batch normalization\cite{ioffe2015batch} and ReLU activation for every intermediate layer, 0.5 of dropout for hidden fully connected layers and stochastic gradient descent with 0.9 Nesterov momentum. Also, we conducted the input normalization simply by dividing standard deviation after subtracting mean value of entire input data. We used Keras\cite{chollet2015} and Theano\cite{al2016theano} framework running on GPUs. Training of local models with small input size such as 18 frames model on MSD have taken about 5 days in total.


\subsection{Compared models}
A typical approach for music auto-tagging is to take about 3 second-long  audio segments as input and average the outputs to make final predictions for an audio clip (e.g. \cite{dieleman2014end}). Here we call them ``local'' models as already denoted in Section \ref{Local feature learning}. On the other hand, we call our proposed architecture ``global'' models as it aggregates features from local models and makes final predictions directly from the audio clip. In our experiment, we evaluate the two models with various combinations of input sizes and feature levels. 

\section{Results and Discussion}
\label{sec:exp}

\subsection{Comparison of local and global models}
Figure \ref{fig:fig3} shows the evaluation results for the local and global models on MTAT and MSD for different input sizes. From the local models, the AUC reaches the highest level when the input size is 108 frames (about 2.5 second), indicating that taking 3 second as input size is actually a reasonable choice. However, the proposed global models consistently outperform the local models and the performance increment is more vivid on MSD. 

\begin{figure*}[t]
\vspace{-5mm}
\centering
\centerline{\includegraphics[width=19cm]{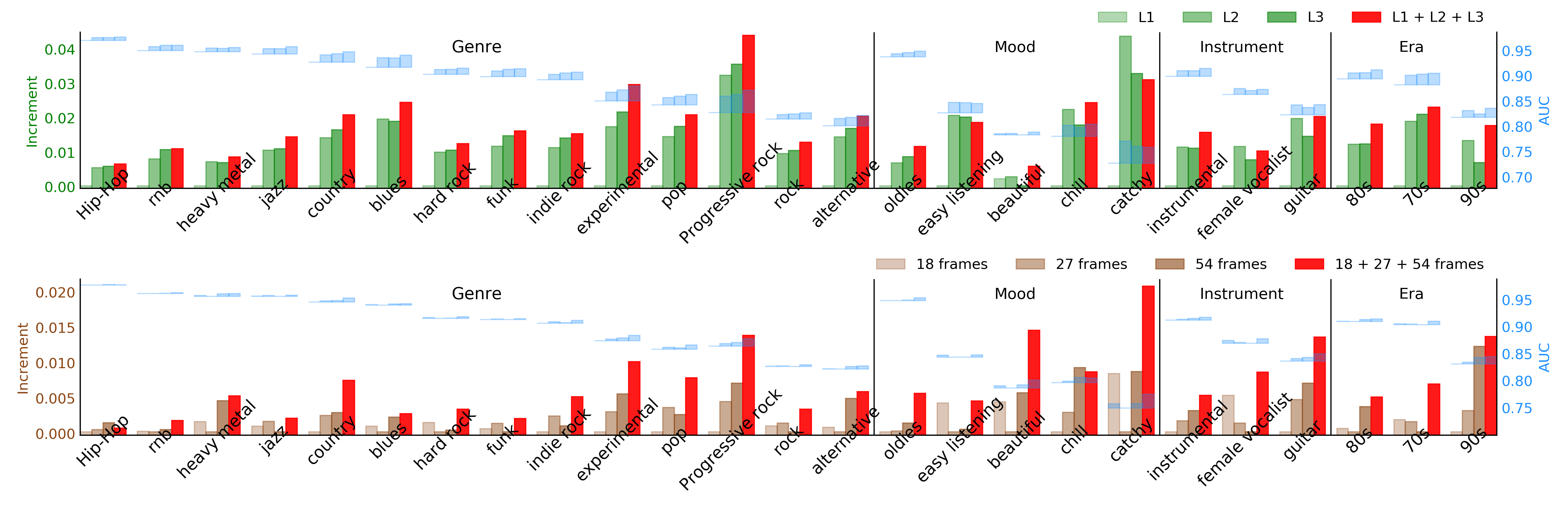}}
\caption{AUC results of the global model on MSD for selected tags. The top compares multi-level features (L1, L2, L3) and their concatenation when the input size is 27 frames. The bottom compares multi-scale features (18, 27, 54 frames) and their concatenation. The right blue bar-graphs scaled by the y-axis on the right side represents the absolute AUCs by which the tags are sorted. The left green (top) and brown (bottom) bar-graph scaled by the y-axis on the left-side represents the performance increments given the the smallest level on individual tags.}
\label{fig:fig6}
\end{figure*}


\subsection{Effect of multi-level features}
Figure \ref{fig:fig4} dissects the effect of multi-level features further in the global model. When a single-level feature is used, higher-level features (L3) apparently work better than lower ones (L1 and L2). When multi-level features are concatenated, the AUC levels consistently increases on both datasets. One interesting result is that each layer have different importance. For example, on MTAT, the absence of L1 features decreases the AUC most. On the other hand, on MSD, the absence of L3 features makes a highest drop. This may attribute to difference in tag words between the two datasets. For example, MTAT contains more instrument-related tags than genre or mood tags, compared to MSD. 


\subsection{Effect of multi-scaled features}
We now discuss the effect of multi-scale features. Figure \ref{fig:fig5} shows the results for different combinations of varying input size in the global models. Compared to multi-level features, the performance gain is not strong but the use of multi-scaled features are definitely helpful. The best result is achieved in both MSD and MTAT when 18, 27 and 54 frames models are combined. This may be inferred from Figure \ref{fig:fig3}. 

\subsection{Performances visualization of individual tags }
We investigate the global model even further by comparing the performance sensitivity of individual tags to different feature levels and time scales as illustrated in Figure \ref{fig:fig6}. In the multi-level comparison (top), since supervised training is performed with the tags in the local feature learning stage, gradual increment is expected as the the layer goes up. This trend, however, does not work consistently for every single tag. For example, some tags including \textit{blues}, \textit{chill}, \textit{guitar} and \textit{80s} favor L2 features more than L3. The non-gradual increment is observed in the multi-scale comparison (bottom) as well. Some tags including \textit{heavy metal}, \textit{experimental}, \textit{progressive rock}, \textit{alternative}, \textit{chill}, \textit{guitar} and \textit{90s} favor 54 frames whereas others including \textit{hard rock}, \textit{easy listening}, \textit{female vocalist} and \textit{70s} work better on 18 frames. Overall, we can ensure that the best AUCs in almost all tags are achieved when the multi-level and multi-scale features are concatenated.



\subsection{Transfer learning and comparison to state-of-the-arts}
In Table \ref{table:table2}, we show the performance of the proposed architecture in the transfer learning settings where MSD is used to pre-train the CNNs as a feature extractor and other datasets are for the final classification. Interestingly, the auto-tagging performance on MTAT is even greater than those using local models trained from the MTAT dataset itself. Also, it shows the music genre classification results on fault-filtered GTZAN and Tagtraum genre annotations on MSD. To our knowledge, we report the performance on the Tagtraum for the first time. From Table \ref{table:table3}, the accuracy on the fault-filtered GTZAN is greater than previously reported ones. Table \ref{table:table3} also compares the best results in the local and global models to those from previous state-of-the-arts on MTAT and MSD. They show that our proposed architecture is highly effective. 


\begin{table}[t]
\vspace{-3mm}
\centering
\caption{Global classification results using features extracted from the CNN pre-trained with MSD}
\label{table:table2}
\begin{tabular}{@{}llll@{}}
\toprule
Dataset  & Task                 & Measurement &                 \\ \midrule
MTAT      & Auto-tagging         & AUC         & 0.9021 \\
Tagtraum & Genre classification & Accuracy    & 0.766           \\ 
GTZAN (fault-filtered)   & Genre classification & Accuracy    & 0.720           \\ \bottomrule
\end{tabular}
\end{table}

\begin{table}[t]
\vspace{-0mm}
\centering
\caption{Comparison of our models to prior state-of-the-arts}
\label{table:table3}
\begin{tabular}{@{}cccc@{}}
\toprule
Model                                                                                                  & MTAT    & MSD    & \begin{tabular}[c]{@{}c@{}}GTZAN\\ (fault-filtered)\end{tabular}  \\ \midrule
Bag of multi-scaled features {\cite{dieleman2013multiscale}}                                          & 0.898  & -      & -     \\
1D CNN {\cite{dieleman2014end}}                                                  & 0.8815 & -      & -     \\
Transfer learning {\cite{van2014transfer}}                                                  & 0.88 & -      & -     \\
Persistent CNN {\cite{liu2016applying}}                                                              & 0.9013 & -      & -     \\
Time-Frequency CNN {\cite{gucclu2016brains}}                                                              & 0.9007 & -      & -     \\
2D CNN {\cite{choi2016automatic}}                                                           & 0.894  & 0.851  & -     \\
CRNN {\cite{choi2016convolutional}}                                                       & -      & 0.862  & -     \\
2D CNN {\cite{kereliuk2015deep}}                                                                                       & -      & -      & 0.632 \\
Temporal features {\cite{jeonglearning}}                                                                                          & -      & -      & 0.659 \\ \midrule
Local model                                                                                            & 0.8981 & 0.8783 & -     \\
Global model with multi-level features                                                                 & 0.9002 & 0.8853 & -     \\
Global model with both multi features                                                                  & 0.9017 & \textbf{0.8878} & -     \\
\begin{tabular}[c]{@{}c@{}}Global model with both multi features\\ (pre-trained with MSD)\end{tabular} & \textbf{0.9021} & -      & \textbf{0.720} \\ \bottomrule
\end{tabular}
\end{table}

\section{Conclusion and Future Work}
\label{sec:concl}

In this paper, we presented a CNN-based architecture, which is designed considering different levels of abstractions in music tags. We showed the effectiveness of the architecture by evaluating different combinations of the multi-level and multi-scale features and also by applying it to transfer learning settings. Finally, we showed that our proposed architecture achieves the best results on the three popularly used datasets. As future work, we plan to train the architecture in a multi-task learning manner by optimizing the local CNNs and global aggregated networks simultaneously.  




\section{Acknowledgment}
This work was supported by Korea Advanced Institute of Science and Technology (Project No. G04140049) and the National Research Foundation of Korea (Project No. 2015R1C1A1A02036962).

\vfill\pagebreak

\ifCLASSOPTIONcaptionsoff
  \newpage
\fi



%

\bibliography{bibtex/bib/IEEEabrv.bib,bibtex/bib/IEEEexample.bib}{}
\bibliographystyle{IEEEtran}





\end{document}